\documentclass{article}

\usepackage{times}
\usepackage{graphicx} 
\usepackage{subfigure}

\usepackage{natbib}

\usepackage{algorithm}
\usepackage{algorithmic}

\usepackage{hyperref}

\usepackage[accepted]{icml2014}

 \usepackage{epstopdf}
 \usepackage{appendix}
\usepackage{amsmath,amssymb,enumerate,amsfonts,url,multirow,array,tabularx,booktabs,stfloats,flushend,color}
\usepackage{amsthm}
\usepackage[T1]{fontenc}
\usepackage{enumitem}
\setenumerate[1]{itemsep=0pt,partopsep=0pt,parsep=\parskip,topsep=5pt}
\setitemize[1]{itemsep=0pt,partopsep=0pt,parsep=\parskip,topsep=0pt}
\setdescription{itemsep=0pt,partopsep=0pt,parsep=\parskip,topsep=5pt}

\def\x{{\mathbf x}}
\def\w{{\mathbf w}}

\def\u{{\mathbf u}}

\def\h{{\mathbf h}}

\def\W{{\mathbf W}}

\def\O{{\mathcal O}}

\begin{document}

\twocolumn[
\icmltitle{Learning Deep Representations By Distributed Random Samplings}

\icmlauthor{Xiao-Lei Zhang}{huoshan6@126.com}
\icmladdress{Tsinghua National Laboratory for Information Science and Technology, Tsinghua University, 100084, China}

\icmlkeywords{boring formatting information, machine learning, ICML}

\vskip 0.3in
]

\begin{abstract}
In this paper, we propose an extremely simple deep model for the unsupervised nonlinear dimensionality reduction -- deep distributed random samplings. First, its network structure is novel: each layer of the network is a group of mutually independent $k$-centers clusterings. Second, its learning method is extremely simple: the $k$ centers of each clustering are only $k$ randomly selected examples from the training data; for small-scale data sets, the $k$ centers are further randomly reconstructed by a simple cyclic-shift operation.
Experimental results on nonlinear dimensionality reduction show that the proposed method can learn abstract representations on both large-scale and small-scale problems, and meanwhile is much faster than deep neural networks on large-scale problems.

\end{abstract}

  \setlength{\arraycolsep}{0.0em}
\section{Introduction}

Deep learning is one of the most powerful representation learning techniques. 
 Recent advanced work starts from Hinton and Salakhutdinov \cite{hinton2006reducing}. The method, named deep belief networks (DBN), contains two phases -- the bottom-up greedy layer-wise pretraining phase and the top-down fine-tuning phase.


In this paper, we propose a simple deep learning method -- deep distributed random samplings (DDRS), for the unsupervised dimensionality reduction. 
The time and storage complexities of DDRS scale linearly with the size of the data set.


In the remainder of the paper, we will describe DDRS in Section \ref{sec:alg}, and report the experimental results in Section \ref{sec:experiment}. Finally, we conclude this paper in Section \ref{sec:conclu}. The theoretical justification, motivation, related work and several supplemental experiments are in the supplementary material.





 \section{Algorithm description}\label{sec:alg}

 \begin{figure}[t]
\center
\scalebox{1}{
\centerline{\includegraphics[width=6cm]{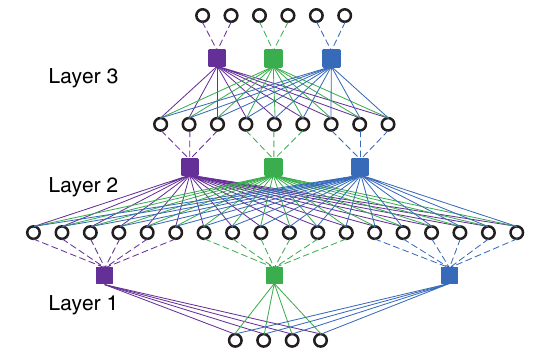}}}
\caption{{Architecture of the proposed DDRS. The clusterings in the same layer are drawn in different colors. Each clustering contains two computation nodes. The first node is the $k$-randomly-sampled examples, which is represented as a square. The second node is the one-hot sparse encoding, which is represented as the dotted lines from the square.} }
\label{fig:softoutput}
\end{figure}

 DDRS is trained layer-wisely (Figure \ref{fig:softoutput}). The output of one layer is the input of its upper layer.
 Each layer consists of $V$ independent $k$-centers clusterings. Given a set of $d$-dimensional input data $\mathcal{X} = \left\{\x_1,\ldots,\x_n \right\}$, the training process of the $v$-th clustering is as follows:

\begin{enumerate}[noitemsep]
\vspace{-1.5mm}
\item  \textbf{Random feature selection:} randomly select $\lfloor ad\rfloor$ dimensions of $\mathcal{X}$ to form a subset of $\mathcal{X}$, denoted as $\mathcal{X}^{(v)} = \left\{\x^{(v)}_1,\ldots,\x^{(v)}_n \right\}$, where $a$ represents the fraction of the selected dimensions.

  \item \textbf{Random sampling:} randomly select $k$ examples from $\mathcal{X}^{(v)}$, denoted as $\W_v = [\w_{v,1},\ldots,\w_{v,k} ]$, as the $k$ centers of the $v$-th clustering.

        \begin{figure}[t]
\center
\scalebox{1}{
\centerline{\includegraphics[width=6cm]{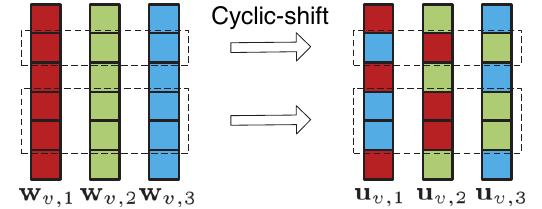}}}
\caption{{An example of the cyclic-shift operation. $\{\w_{v,1},\w_{v,2},\w_{v,3}\}$} are three centers of the $v$-th clustering that are randomly selected from the training data. $\{\u_{v,1},\u_{v,2},\u_{v,3}\}$ are the centers that have been handled by the cyclic-shift operation. The squares represent the entries of the vectors.}
\label{fig:cyclic}
\end{figure}

  \item \textbf{Random reconstruction:} Randomly select $\lfloor r\lfloor ad\rfloor \rfloor$ dimensions of $\W_v$, and do one-step cyclic-shift on the selected dimensions as in Figure \ref{fig:cyclic}, where $r$ represents the fraction of the randomly selected features over the $\lfloor ad\rfloor$-dimensional centers.

  \item \textbf{Sparse representation learning:} (i) calculate the similarities $\h_{v}$ between the input $\x^{(v)}$ and the $k$ centers in terms of some predefined similarity measurement at the bottom layer, such as the Euclidean distance, and in terms of $\h_{v} = \W_v^T\x^{(v)}$ at all other layers; (ii) enforce one-hot encoding on $\h_v$ by setting the entry that corresponds to the closest center to $\x^{(v)}$ to 1 and all other entries to 0. 
 \end{enumerate}
\vspace{-1.5mm}

The outputs of all clusterings of each layer are concatenated to a long sparse vector, i.e. $\bar{\h} = [\h_1^T,\ldots,\h_V^T]^T$.



 \section{Empirical evaluation}\label{sec:experiment}
In this section, we will focus on the unsupervised dimensionality reduction problem. When we evaluate the running time, the experiments are run with \textit{one}-core PC with 8 GB memory. The experiments are conduced on three data sets, which are the MNIST handwritten digits, and a small-scale data set respectively. The analysis on how the parameters affect the performance is attached in the supplementary material.

The bottom layer of DDRS uses the linear kernel to calculate the similarities between the input data and the centers in all data sets. Because DDRS learns only a sparse high-dimensional representation, we use principle component analysis (PCA) to project it to a low-dimensional subspace. Only a few largest eigenvalues and their corresponding eigenvectors are preserved for constructing the subspace.

\subsection{Results on the MNIST digits}

MNIST handwritten digit data set is a benchmark data set that contains 10 hand written integer digits ranging from 0 to 9. It consists of 60,000 training examples and 10,000 test examples. Each example has 784 dimensions. We normalize each example to $[0,1]$ by dividing each entry of the example by 255.

The parameter setting of DDRS is as follows. The learned representations are projected to $\{2,3,5,10,20,30\}$ low-dimensional subspaces respectively.

The parameter settings of DBN are the same as in \cite{hinton2006reducing} except that the number of the units in the linear output layer are set to $\{2,3,5,10,20,30\}$ respectively. The CPU time consumed on pretraining and fine-tuning are recorded separately.

 \begin{figure}[t]
\center
         \includegraphics[width=1cm]{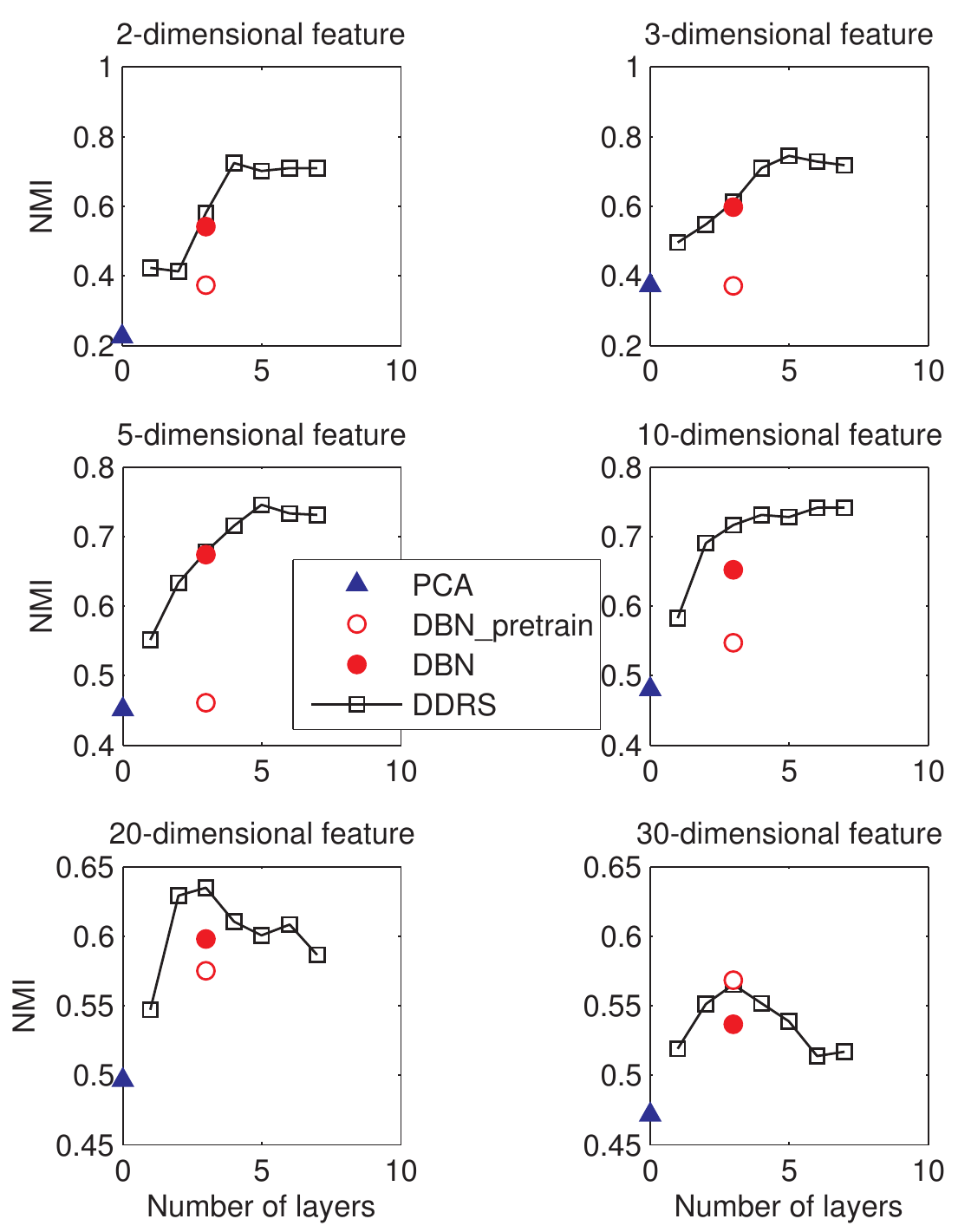}
 \caption{{Normalized mutual information (NMI) comparison of the $k$-means clusterings.} }
 \label{fig:MNIST_feature_kmeans}
 \end{figure}

  \begin{figure}[t]
\center
         \includegraphics[width=1cm]{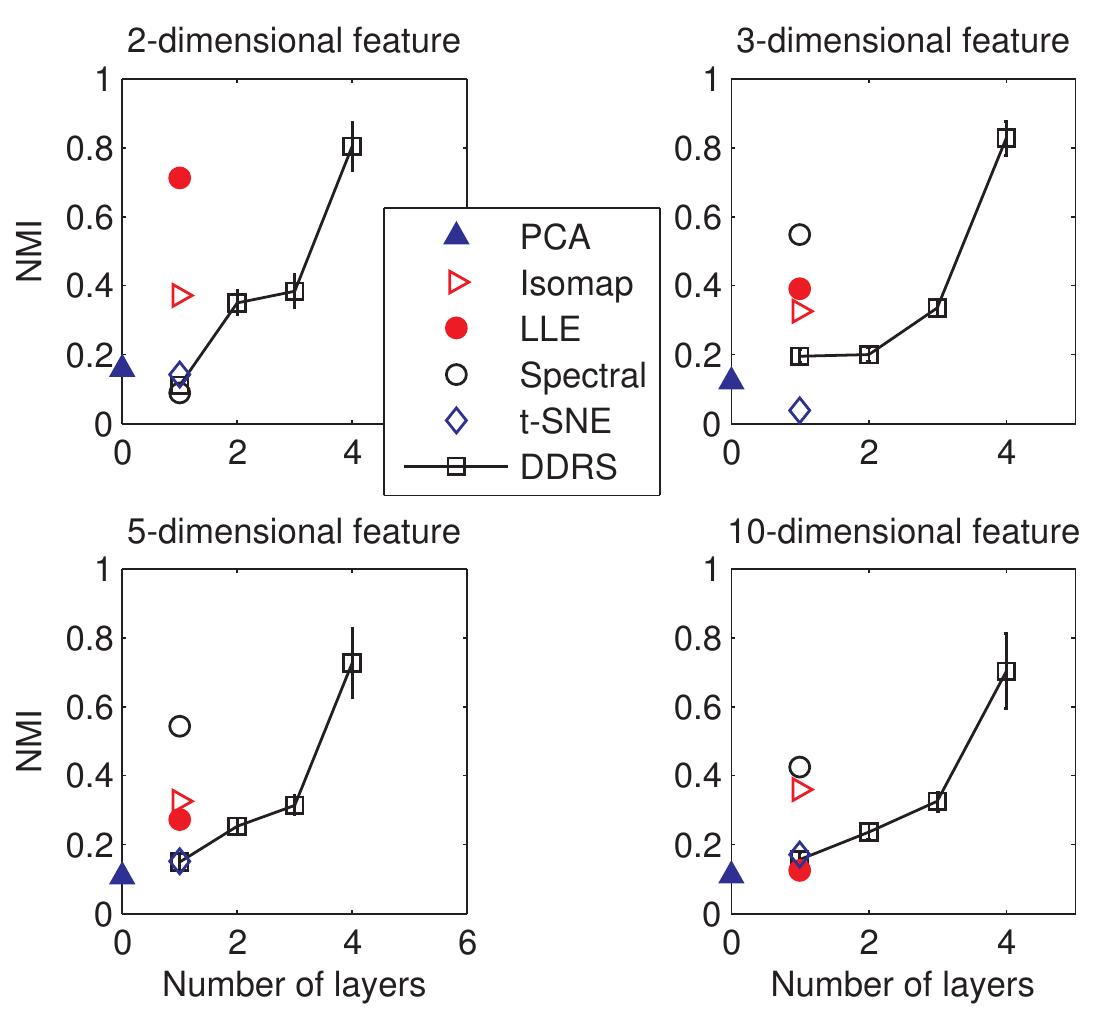}
 \caption{{NMI comparisons of the $k$-means clusterings using the features learned by the competitive methods on the small-scale data set.} }
 \label{fig:ALLAML_acc_kmeans}
 \end{figure}

We use the training set to train the models, and evaluate the effectiveness of the learned representations on the test set by the $k$-means clustering, where the Euclidean distance is used to measure the similarity between any two examples in the low-dimensional subspace. The \textit{normalized mutual information} (NMI) \cite{strehl2003cluster} is used as the evaluation metric, and the results are average ones over 10 independent runs. We will compare DDRS with DBN and PCA.



The results of using the $k$-means clustering are summarized in Figures \ref{fig:MNIST_feature_kmeans}. From the figures, we can see that (i) when the dimensions are restricted to 2 to 5, DDRS is as good as DBN as long as they are in the same depth, and significantly outperforms PCA; (ii) when the dimensions are enlarged to 10 to 30, DDRS performs better than PCA and DBN; (iii) when the dimensions are gradually increased, all methods are getting worse and worse.

The CPU time is recorded in Table \ref{tab:contentCompare1}. From the table, we can see that DDRS is about 20 times faster than DBN.

We also compared DDRS with other well-known dimensionality reduction methods on small subsets of MNIST in the supplementary material, since all of the supplemental competitive methods cannot handle large scale problems. DDRS is at least as good as the best competitors.

\begin{table}[t]
\caption{\label{tab:contentCompare1} {CPU time (in hours) comparison on MNIST.}}
\centerline{
\scalebox{0.03}{
\begin{tabular}{l||c|c|c}
 \hline
& DBN$_{\scriptsize\mbox{pretraining}}$ & DBN$_{\scriptsize\mbox{fine\_tunning}}$& DDRS \\
 \hline\hline
\textbf{Time}&2.94 &74.54 & 5.16\\
\hline
\end{tabular}}}
\end{table}


\subsection{Results on a small-scale data set}

Sometimes, we have to deal with very small-scale but high-dimensional data sets.
In this subsection, we will study such a very small-scale problem. The data set is a two-class classification problem that consists of 38 training examples and 34 test examples.

The parameter settings of DDRS is as follows. The learned representations are projected to $\{2,3,5,10\}$ low-dimensional subspaces.

We will compare DDRS with PCA and some methods based on graphs. The competitive methods are all one-layer nonlinear dimension reduction methods that have an $\O(n^2)$ complexity.

We run the experiment 10 times and report the average performance.
When $k$-means clustering is used for evaluation, for each single experimental running, we run $k$-means on the entire data set 50 times and record the average NMI performance.
Figure \ref{fig:ALLAML_acc_kmeans} shows the performance of the $k$-means clusterings using the features learned by the competitive methods. From the figures, we observe that DDRS can learn a feature that is at least as good as the best competitor, and the experimental phenomena is similar with those on the MNIST data set.

\section{Conclusions}\label{sec:conclu}

DDRS is simple, fast, and effective.

\section*{Acknowledgement}
The author would like to thank Prof. DeLiang Wang for providing the Ohio Supercomputing Center, Columbus, OH, USA, for the experimental running.

\bibliography{mywork,zxlrefs}
\bibliographystyle{icml2014}

\end{document}